\documentclass[letterpaper]{article} 
\usepackage{aaai2026}  

\usepackage{times}  
\usepackage{helvet}  
\usepackage{courier}  
\usepackage[hyphens]{url}  
\usepackage{graphicx} 
\usepackage{natbib}  
\usepackage{caption} 
\usepackage{algorithm}
\usepackage{algorithmic}
\usepackage{newfloat}
\usepackage{listings}
\usepackage{booktabs}
\usepackage{xcolor}         

\usepackage{xcolor,colortbl}

\usepackage{subcaption}

\DeclareCaptionStyle{ruled}{labelfont=normalfont,labelsep=colon,strut=off} 
\floatstyle{ruled}
\newfloat{listing}{tb}{lst}{}
\floatname{listing}{Listing}
\nocopyright

\newcommand{\DatasetName}{TextGround4M\xspace}

\usepackage{amsmath}
\usepackage{xspace}

\title{TextGround4M: A Prompt-Aligned Dataset for Layout-Aware Text Rendering}
\author {
    Dongxing Mao\textsuperscript{\rm 1}, 
    Yilin Wang\textsuperscript{\rm 2}, 
    Linjie Li\textsuperscript{\rm 3}, 
    Zhengyuan Yang\textsuperscript{\rm 3}, 
    Alex Jinpeng Wang\textsuperscript{\rm 1}\thanks{Corresponding author.}
}
\affiliations {
    \textsuperscript{\rm 1}Central South University\\
    \textsuperscript{\rm 2}Zhejiang University\\
    \textsuperscript{\rm 3}Microsoft Research \\
    dongxing.mao@u.nus.edu, jinpengwang@csu.edu.cn
} 

\begin{document}

\maketitle

\begin{abstract}
Despite recent advances in text-to-image (T2I) generation, models still struggle to accurately render prompt-specified text with correct spatial layout—especially in multi-span, structured settings. 
This challenge is driven not only by the lack of datasets that align prompts with the exact text and layout expected in the image, but also by the absence of effective metrics for evaluating layout quality.
To address these issues,  we introduce \DatasetName, \textit{a large-scale dataset} of over 4 million prompt-image pairs, each annotated with span-level text grounded in the prompt and corresponding bounding boxes.
This enables fine-grained supervision for layout-aware, prompt-grounded text rendering.
Building on this, we propose \textit{a lightweight training strategy} for autoregressive T2I models that appends layout-aware span tokens during training, without altering model architecture or inference behavior.
We further construct a \textit{benchmark with stratified layout complexity} to evaluate both open-source and proprietary models in a zero-shot setting. 
In addition, we introduce \textit{two layout-aware metrics} to address the long-standing lack of spatial evaluation in text rendering.
Our results show that models trained on \DatasetName outperform strong baselines in text fidelity, spatial accuracy, and prompt consistency, highlighting the importance of fine-grained layout supervision for grounded T2I generation.

\end{abstract}

\section{Introduction}

Text-to-image (T2I) generation has achieved remarkable progress in recent years, with large-scale generative models such as DALL·E3~\cite{dalle3}, LlamaGen~\cite{llamagen} and Chameleon~\cite{chameleon} demonstrating strong capabilities in synthesizing realistic images from natural language prompts. 
These models, powered by innovations in diffusion and autoregressive (AR) architectures, can produce a stunning diversity of visual content. 
However, a key subtask remains underexplored: prompt-grounded, layout-aware text rendering—accurately generating and positioning the text within the image.

This capability is particularly important for applications such as poster design, digital advertisements, and educational content generation, where \textit{precise textual fidelity and spatial control are essential}.
For example, given a prompt ``a flyer that says 'Summer Sale, 50\% OFF, Aug 12–15' in the top-left corner,'' the model is expected to generate an image containing all specified text content at the indicated location. 
Despite their impressive visual quality, existing models often fail to render text correctly, omit information, or misplace it, limiting their practical usability in such scenarios.

These issues arise from the lack of explicit alignment between the prompt, rendered text, and its spatial position in training data. 
Datasets like AnyWord-3M~\cite{anytext} and MARIO-10M~\cite{textdiffuser} provide OCR annotations but typically label text at the word or line level without connecting these regions to the prompt, leading to misalignment and hindering precise rendering and placement.

\begin{figure}[t]
    \centering
    \includegraphics[width=\linewidth]{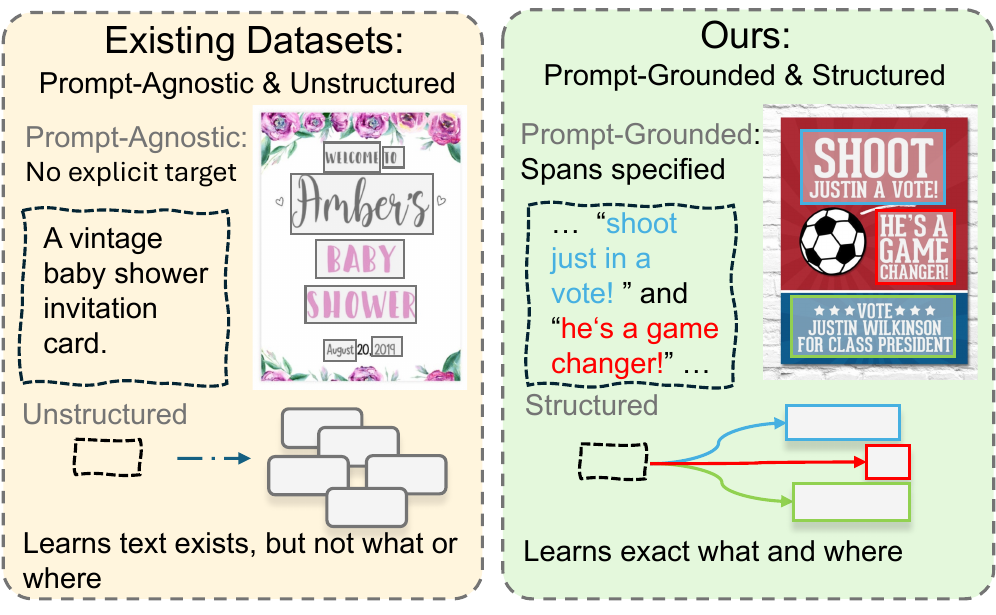}
    \caption{
    Comparison of existing datasets and ours.
    Existing datasets annotate all visible text without prompt grounding or layout structure. Ours offers prompt-aligned, span-level annotations with precise spatial layout, better supporting faithful text rendering.} 
    \label{fig:intro}
\end{figure}
To address this foundational gap, we first present \DatasetName, a \textbf{large-scale, high-quality dataset specifically designed for prompt-grounded text rendering}. 
The comparison against previous datasets is shown in Figure~\ref{fig:intro}.
Each image is paired with a natural language prompt, and all rendered text spans that are semantically entailed by the prompt are explicitly annotated with their textual content and spatial bounding boxes. To construct this dataset at scale and with high fidelity, we design a systematic data pipeline that combines vision-language models (GPT-4o, Qwen2.5-VL) with rule-based alignment and filtering strategies, enabling accurate identification of prompt-grounded text spans and their layouts across millions of samples. This structure provides direct supervision for both content fidelity and layout consistency during training.

Building upon this dataset, we further propose a \textbf{lightweight, training-time-only supervision strategy tailored for autoregressive T2I models}. 
During training, we append structured tokens—including the rendered phrases and their corresponding spatial coordinates—after the image tokens as autoregressive targets. 
This allows the model to leverage layout information as implicit supervision, enhancing alignment learning without requiring any architectural modification or additional inputs at inference time. 

Finally, to enable more rigorous evaluation, we design a difficulty-aware benchmark by stratifying the dataset into levels based on the number of text spans and prompt complexity. 
Using this benchmark, we conduct a comprehensive evaluation of both open-source and commercial models, revealing substantial gaps in their ability to align text content with prompts and render it in visually coherent ways.
To provide more comparison aspect, we also design two new metrics for text rendering evaluation.
Building on these, we validate our approach through experiments on Janus-Pro~\cite{januspro}, demonstrating the effectiveness of our layout-aware, high-quality text rendering strategy without compromising model performance or inference efficiency.

In summary, our contributions are:

\begin{itemize}
    \item We introduce \DatasetName, a large-scale dataset of 4 million prompt-image pairs explicitly annotated with fine-grained, prompt-grounded text spans and their spatial locations. 
    The dataset is constructed using a scalable pipeline combining vision-language models (GPT-4o, Qwen2.5-VL) and rule-based filtering, ensuring \textit{high-quality alignment between prompt instructions and rendered text content}.

    \item We introduce an innovative Layout-posting approach, where the model first predicts images  before generating the layout. 
    This method achieves performance improvements comparable to pre-layout, while \textit{providing flexible and accurate layout rendering,  with free-form inputs}.

    \item  We construct a difficulty-aware benchmark by stratifying samples based on text span count and prompt complexity. We evaluate multiple open-source and commercial models under this unified protocol, revealing significant limitations in prompt fidelity and layout precision, and demonstrating the effectiveness of our approach.

    \item We propose two novel layout-aware metrics for text rendering evaluation, \textit{addressing the lack of standardized assessment for spatial fidelity in text rendering. }

\end{itemize}

\section{Related Works}

\paragraph{Text-Image Datasets for Generation.}
paragraph{Text-Image Datasets for Generation.}
Classic datasets such as MS-COCO~\cite{mscoco} and TextCaps~\cite{textcaps} offer descriptive captions but lack fine-grained alignment between text and its spatial layout. Larger collections like CC3M~\cite{cc12m} and LAION~\cite{laion} increase diversity but mainly provide short, unstructured captions, offering limited support for text rendering.
Recent text-heavy datasets, including MARIO-10M~\cite{textdiffuser} and AnyWords3M~\cite{anytext}, improve textual density but still fall short in prompt-level grounding or layout consistency.

In contrast, our dataset \DatasetName provides span-level, prompt-grounded text annotations with precise spatial layouts, enabling fine-grained supervision for layout-aware T2I generation.

\paragraph{Text-to-Image Generation.}
Recent advances in text-to-image (T2I) generation have been largely driven by diffusion-based models~\cite{llama2,chen2024pixart,sd3, xie2025sana15efficientscaling,chang2023musetexttoimagegenerationmasked} and autoregressive approaches~\cite{llamagen, lumina_mgpt, xin2025luminamgpt20standaloneautoregressive,janus,geng2025xomnireinforcementlearningmakes}. 
These models have demonstrated remarkable capability in synthesizing high-quality and semantically relevant images given natural language prompts. 
Despite their impressive visual fidelity, most existing T2I models struggle with accurately rendering embedded text, particularly in layout-sensitive or prompt-grounded scenarios.

Efforts ~\cite{anytext,textdiffuser, textdiffuser2,du2025textcrafteraccuratelyrenderingmultiple} have explored layout-aware or retrieval-augmented generation to improve controllability. Nevertheless, these models often rely on additional layout inputs or templates during inference, which limits their applicability in free-form generation settings. Our work differs by enabling accurate and grounded text rendering under prompt-only supervision, without explicit layout hints.

\paragraph{Layout-Guided Text Rendering.}
Layout-aware text-to-image generation aims to control not only the textual content but also its spatial placement. This is critical for applications like poster design and instructional diagrams, where text layout affects both function and appearance.
Prior works~\cite{anytext,textdiffuser2, glyphcontrol,glyphdraw} have explored explicit layout conditioning via bounding boxes or templates. Beyond text-specific systems, general layout-conditioned models~\cite{layoutdm,controlnet,plangen, composer,gligen}—demonstrate the value of injecting spatial priors into generative models. However, these approaches often require layout hints at inference time, limiting generalization.
In contrast, our method enables layout-aware rendering under prompt-only supervision, achieved by aligning prompt spans with OCR-detected regions during training.

\section{Dataset Construction}

\begin{figure*}[t]
    \centering
    \includegraphics[width=.9\linewidth]{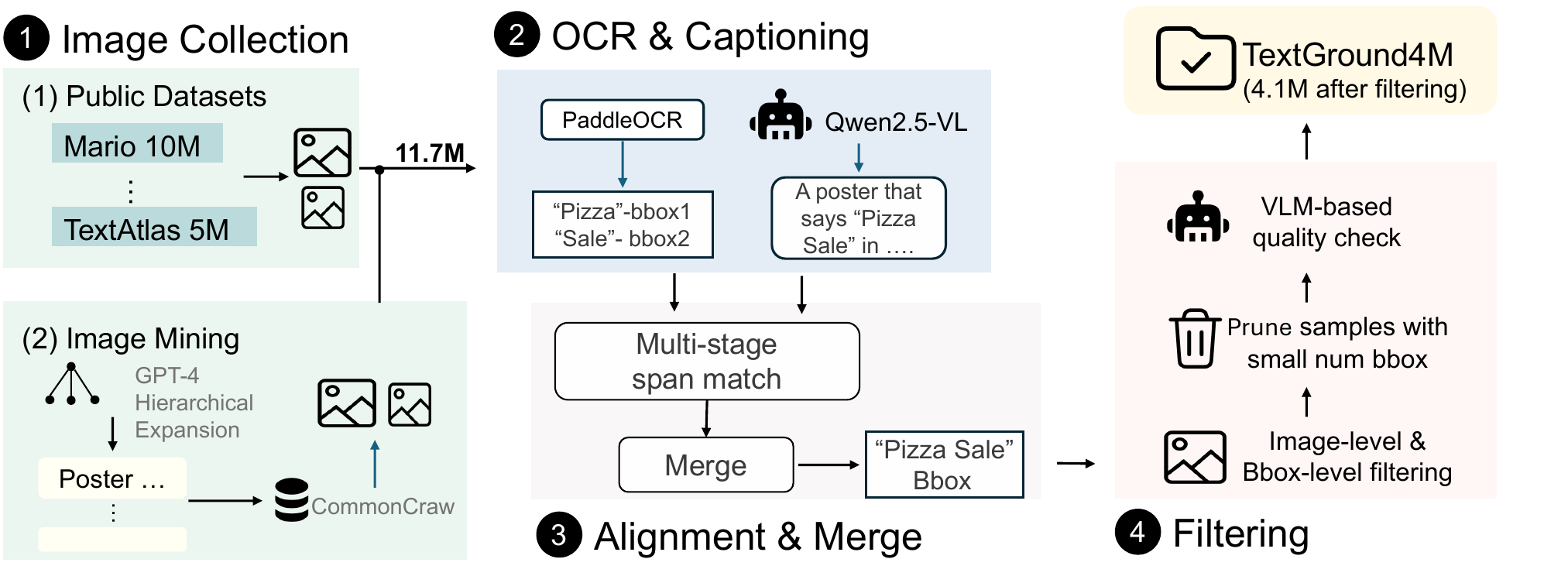}
    \caption{
    Overview of the dataset construction pipeline for TextGround4M.
    We collect 11.7M image-text pairs from both public datasets and GPT-4o-driven Image mining, apply OCR and VLM captioning, perform span-level alignment and merging, and apply multi-stage filtering to obtain 4.1M high-quality samples.} 
    \label{fig:data_construction}
\end{figure*} 

To support fine-grained, prompt-grounded text rendering in T2I models, we construct \DatasetName, a large-scale dataset of 4.1M image-text pairs. Our data construction pipeline combines rule-based heuristics and VLM guidance to ensure high-quality, semantically aligned annotations. Specifically, we integrate public dataset curation, hierarchical image mining, VLM-based captioning, OCR extraction, multistage alignment, and multi-stage quality filtering.

Each image is annotated with bounding boxes that are semantically entailed by the prompt, enabling layout-aware training. The full construction process consists of four stages: image collection, OCR \& captioning, alignment \& merging, and quality filtering, as illustrated in Figure~\ref{fig:data_construction}.

\subsection{Image Collection} 

We collect a total of 11.7M raw image-text pairs from two complementary sources: (1) 8M curated samples from public datasets, and (2) 3.7M samples mining from CommonCrawl\footnote{\url{https://commoncrawl.org/}} using a hierarchically structured retrieval pipeline.

First, we aggregate 8M samples from ten high-quality public datasets, including Mario10M, Anyword3M, and the TextScenesHQ~\cite{wang2025textatlas5mlargescaledatasetdense}. These sources offer broad domain coverage but often contain low-density or trivial cases with minimal visible text or overly simple layouts.

Second, to improve coverage of text-rich and structurally diverse scenarios that are underrepresented in public datasets, we construct a large-scale image-mining subset. We design a hierarchical query generation pipeline powered by GPT-4o, which expands a manually defined domain–topic–subtopic taxonomy into rich, context-aware search phrases. For each subtopic, GPT-4o generates key objects, contexts, and query modifiers that are composed into diverse queries such as 'Highway exit sign on a quiet suburban road' or `Instructional banner with red text at the top.'

These generated queries are used to guide large-scale image mining from CommonCrawl. Specifically, we treat each query as a semantic anchor to retrieve images whose surrounding textual context or OCR-recognized content matches the query intent. Retrieved candidates are then filtered using heuristic criteria to retain only high-resolution, text-heavy images with clear textual regions. We then merge the mined and curated subsets and passed to the next stage.

\subsection{OCR Processing and Prompt Alignment}

To obtain precise supervision, we adopt a hybrid pipeline that combines a vision language model (VLM) and an OCR system, as illustrated in Figure~\ref{fig:data_construction}. Given an image, we first apply Qwen2.5-VL to generate a fine-grained caption, where all visible textual spans are enclosed in quotation marks .

Simultaneously, we use PaddleOCR~\cite{cui2025paddleocr30technicalreport,cui2025paddleocrvlboostingmultilingualdocument} to extract word-level text (e.g. ``Pizza'', ``Sale'') and their bounding boxes. Since the VLM outputs sentence-level descriptions and the OCR yields discrete tokens, we introduce a multistage alignment module to associate quoted spans in the caption with detected text boxes. The alignment is based on heuristic rules, including exact match, partial overlap, and fuzzy string matching.
This process produces training annotations that are both spatially grounded and semantically aligned with the prompt. 
See appendix for alignment algorithm details.

\subsection{Quality Filtering}

To ensure high-quality annotations, we apply a three-stage filtering pipeline, as shown in Figure~\ref{fig:data_construction}.

\textbf{(1) Image-level and BBox-level Filtering.}  
We begin by removing low-quality or irrelevant samples using heuristic rules. Following Anyword3M, we discard images with width or height below 256 pixels or with aspect ratios outside [0.67, 1.5]. Samples are also excluded if OCR-detected text regions occupy less than 10\% of the image area, or if the OCR confidence score falls below 0.7. Additionally, text lines that are unreadable, empty, or composed entirely of symbols are filtered out.

To ensure semantic alignment, we discard samples in which more than 70\% of OCR spans cannot be matched to any quoted span in the VLM-generated caption. This criterion helps retain only prompt-relevant and meaningful text.

\textbf{(2) Pruning Trivial Cases.}  
To reduce over-representation of trivial examples, we selectively downsample samples with limited text content. Specifically, we discard approximately 60\% of samples with only one bounding box and 40\% of those with exactly two, while keeping a balanced share to maintain distributional robustness.

\textbf{(3) VLM-based Semantic Filtering.}  
As a final quality control step, we re-invoke Qwen2.5-VL to verify the semantic completeness and consistency of each sample. We enforce two conditions: (i) all quoted spans in the caption must be grounded to valid OCR-detected boxes; and (ii) merged annotations must reflect faithful and spatially coherent alignment between vision and language. This stage functions as a semantic auditor that detects misalignment, omission, or over-labeling errors.

After all filtering stages, we retain 4.1M high-quality samples as our final dataset, \DatasetName. Among these, 3.9M are derived from rigorously filtered public datasets and 0.2M from our image-mining subset. Despite their smaller share, image-mining samples \textit{significantly enrich the dataset with complex and underrepresented text scenarios, contributing disproportionately to its semantic diversity}.

\subsection{Dataset Statistics}  

\begin{figure}[t]
    \centering

    \includegraphics[width=\linewidth]{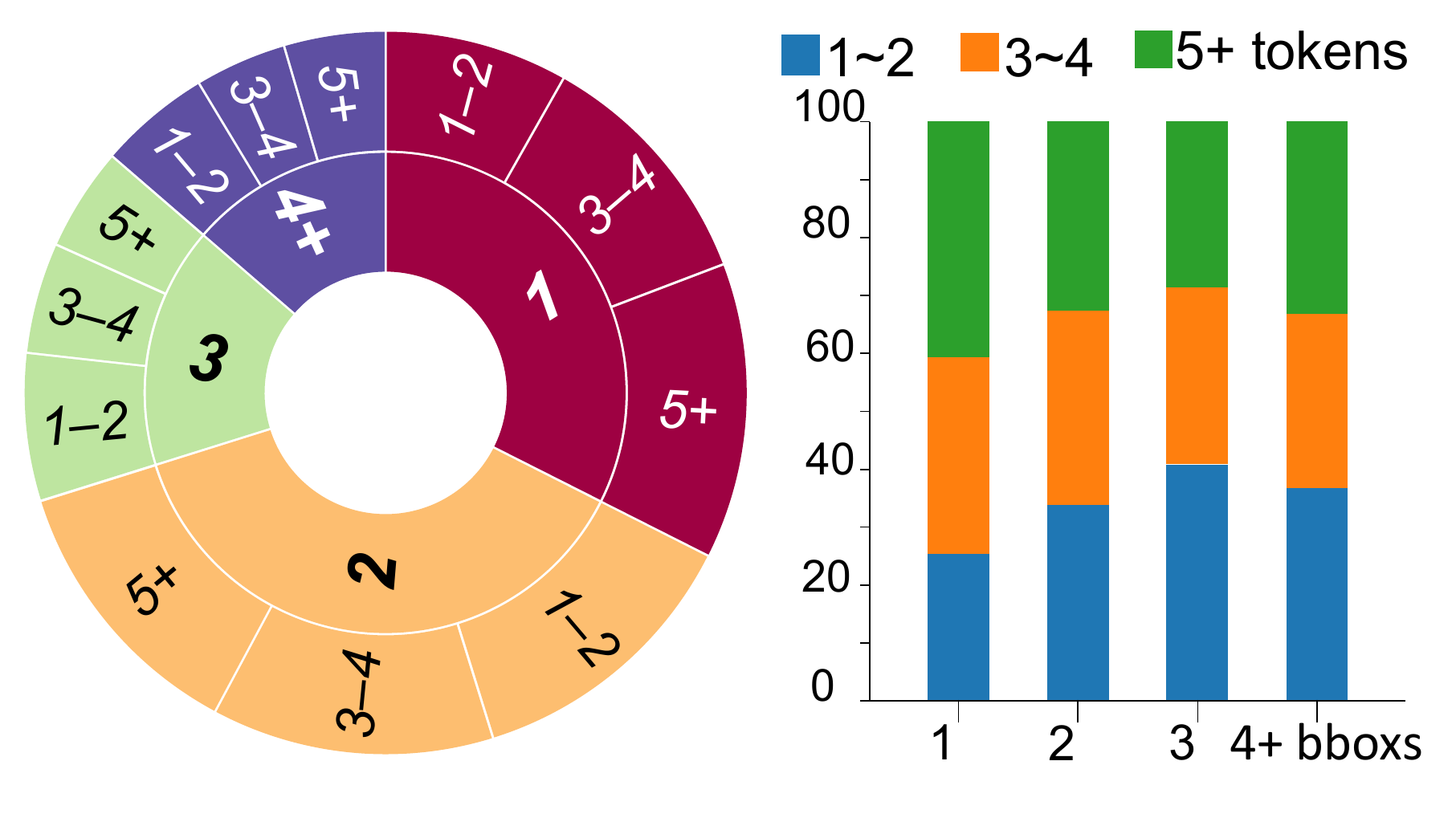}

    \caption{
    Visualization of structural statistics in \DatasetName. 
    (a) The dual-ring pie chart shows the distribution of bounding box counts (inner ring) and per-box token lengths (outer ring). 
    (b) The stacked bar chart reveals token length composition under each bbox group.
    }
    \label{fig:bbox_token_stats}
\end{figure}

\paragraph{Structural Distribution of Text Instances.}
Figure~\ref{fig:bbox_token_stats}(a) shows a nested pie chart illustrating the joint distribution of bbox counts per image (inner ring) and token lengths per bbox (outer ring). 
Overall, images with 1–2 bboxes constitute over 57\% of the dataset, with 1-bbox samples alone accounting for 32.4\%. 
While short text spans (1--4 tokens) dominate these simpler layouts, 13.2\% of 1-bbox samples still contain long text (5+ tokens), indicating that low layout complexity does not necessarily imply low semantic content.

\paragraph{Semantic Topic Coverage.}
To enhance topic diversity, we mined a large number of images from the web using a hierarchical topic structure. Specifically, we organized the mining process into 7 major scenarios (e.g., signage, packaging) and further expanded into 90 distinct topics and over 2,900 fine-grained subtopics.
The final topic distribution of the mined images is provided in the supplementary material.

\section{Method}

\subsection{Overview}

We propose a lightweight training strategy to enhance text rendering quality in autoregressive (AR) text-to-image (T2I) generation. Our key idea is to introduce \textbf{prompt-grounded sentence-level supervision} into the training  by appending the sentence tokens and their corresponding bounding box tokens to the image token sequence as autoregressive targets. 
This enables the model to learn a tighter alignment between prompt semantics and image-level text content, without modifying model architecture or decoding procedures.

\begin{figure*}[t]
    \centering
    \includegraphics[width=.9\linewidth]{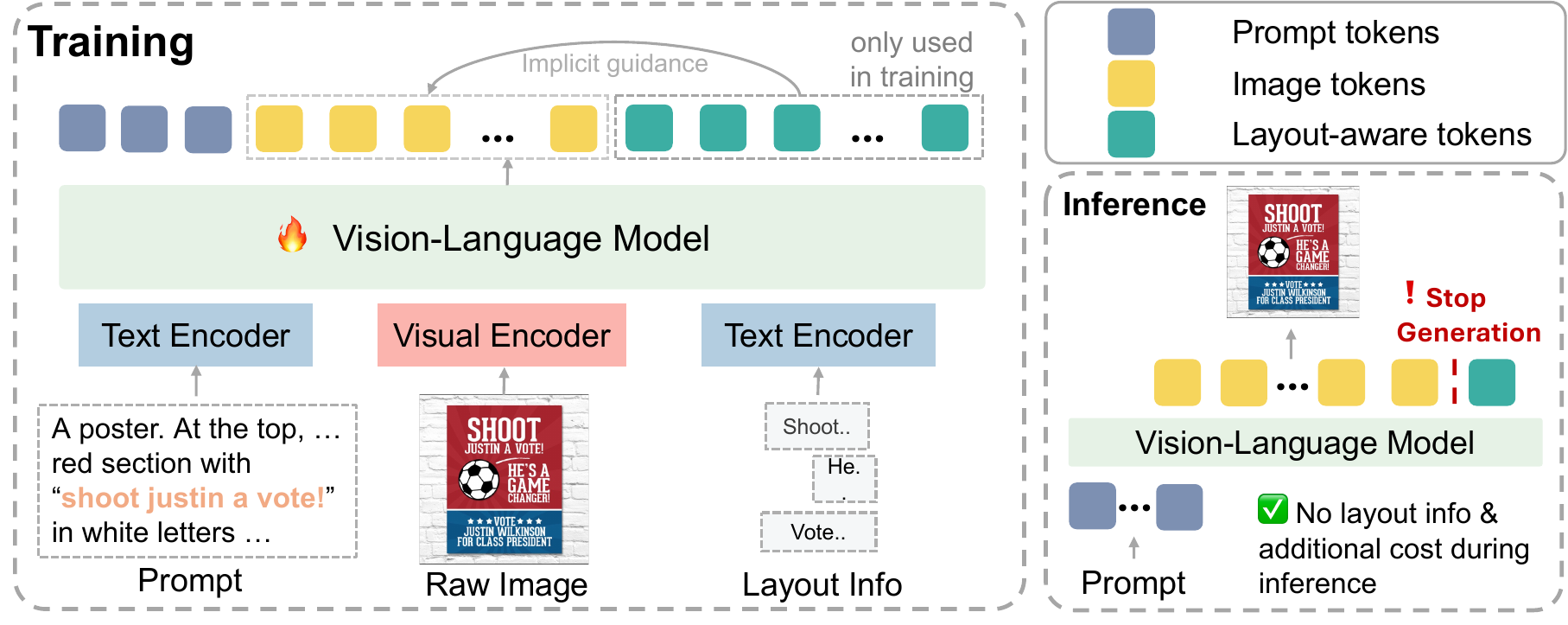}
    \caption{
    Training and inference pipeline of our method.
    During training, prompt-grounded span tokens and bounding box tokens are appended after visual tokens to provide layout-aware supervision. At inference time, the model autoregressively generates image tokens only, without requiring any layout annotations or architectural changes.} 
    \label{fig:method}
\end{figure*}

As shown in Figure~\ref{fig:method}, given a prompt, the model autoregressively generates image tokens, followed by the expected OCR content, including each rendered word and its corresponding bounding box. At inference time, only the image tokens are generated, making the method plug-and-play for any AR-based T2I backbone.

\subsection{Token Sequence Construction}

Let $p$ denote the input prompt and $I$ the corresponding ground-truth image. We assume that $I$ contains $N$ textual regions explicitly required by the prompt, denoted as $\{w_1, \dots, w_N\}$, with each associated bounding box represented as $b_i = (x_i^{\text{min}}, y_i^{\text{min}}, x_i^{\text{max}}, y_i^{\text{max}})$. These coordinates are normalized to integers in the range $[0, 512]$, relative to the image resolution, and specify the upper-left and lower-right corners of each bounding box. Here, ``explicitly required'' refers to text that is semantically entailed by the prompt (e.g., \textit{``a sign that says ‘HELLO WORLD’''}), as opposed to incidental scene text.

The training target sequence $\mathcal{T}$ is constructed as:
\[
\mathcal{T} = \texttt{ImageTokens}(I) \,\|\, \big\|_{i=1}^{N} \left[ \texttt{TextTokens}(w_i,b_i) \right]
\]
where $\texttt{ImageTokens}(I)$ denotes a discrete representation of the image $I$, and each $\texttt{TextTokens}(b_i, w_i)$ jointly encodes the spatial and textual information of the $i$-th prompt-grounded text region. Both the box coordinates and the word contents are tokenized into discrete sequences. This structure enables the model to learn the alignment between visual regions and prompt-driven semantics.
During training, the model autoregressively predicts $\mathcal{T}$ conditioned on the prompt $p$, providing joint supervision for image generation and accurate, prompt-grounded text rendering.

\subsection{Training Objective}

As described in the previous section, the target token sequence $\mathcal{T}$ is constructed by concatenating visual tokens derived from the image and prompt-grounded text tokens that encode the desired textual content and spatial layout. Since these two parts serve distinct learning objectives—image and text—we decouple the supervision accordingly.

As described in the previous section, the target sequence $\mathcal{T}$ is composed of two parts: 
$\mathcal{T}_{\text{img}}$, the visual tokens derived from the image, and 
$\mathcal{T}_{\text{text}}$, the prompt-grounded text tokens encoding desired content and layout. 
Since these segments serve distinct learning purposes—image synthesis and text rendering—we apply 
separate autoregressive objectives:$
\mathcal{L}_{\text{img}}  = - \sum_{t \in \mathcal{T}_{\text{img}}} \log P(t_t \mid t_{<t}, p),$ and $\mathcal{L}_{\text{text}} = - \sum_{t \in \mathcal{T}_{\text{text}}} \log P(t_t \mid t_{<t}, p)$.

The total training loss is a weighted sum:
$\mathcal{L} = \mathcal{L}_{\text{img}} + \alpha \cdot \mathcal{L}_{\text{text}}$,
where $\alpha$ is a tunable hyperparameter that balances the visual and textual supervision. Here we set $\alpha=1$ by default.

This decoupled formulation enables the model to separately optimize for visual token decoding and prompt-aligned text generation, promoting both visual realism and faithful textual grounding.

\subsection{Training-Time Supervision and Inference}

A key design principle of our approach is that \textit{prompt-grounded text supervision is applied exclusively during training}. Specifically, we append structured tokens—encoding the textual content and spatial coordinates of words required by the prompt—after the image tokens in the target sequence. This additional supervision guides the model to associate prompt semantics with the appearance and layout of rendered text in the image.

At inference time, however, these textual tokens are entirely omitted. The model is given only the prompt $p$ and autoregressively generates image tokens, following the standard decoding process of AR-based T2I models.

This training-time-only supervision strategy benefits from:
\emph{i}. \textit{Architectural compatibility:} It requires no modifications to the underlying model architecture or decoding procedure.
\emph{ii}. \textit{Inference simplicity:} No additional modules, layout annotations, or OCR feedback are needed at test time.
\emph{iii}. \textit{Implicit grounding:} The model learns to internalize the correspondence between prompt content and rendered text, enhancing generalization without explicit control.

\section{TextGround-Bench}

To systematically evaluate prompt-grounded text rendering performance, we propose TextGround-Bench, a difficulty-aware benchmark based on our TextGround dataset. This benchmark enables fine-grained comparison across models with varying levels of semantic and spatial complexity.

\subsection{Test Set Construction}

We define difficulty levels using a rule-based strategy based on (i) the number of prompt-grounded bounding boxes ($N_{\text{box}}$), and (ii) the maximum number of tokens within any span ($W_{\text{max}}$). The categories are:
\emph{i}. \textit{Easy}: $N_{\text{box}} \leq 2$ and $W_{\text{max}} \leq 4$.
\emph{ii}. \textit{Medium}: ($N_{\text{box}} \geq 3$ and $W_{\text{max}} \leq 4$) or ($N_{\text{box}} \leq 2$ and $W_{\text{max}} \geq 5$).
\emph{iii}. \textit{Hard}: $N_{\text{box}} \geq 3$ and $W_{\text{max}} \geq 5$.
The rule ensures clear difficulty assignment and distinguishes structural from linguistic complexity. Distribution stats and examples are provided in the supplementary.
\subsection{Evaluation Metrics}

To comprehensively assess the quality of generated images in text-to-image (T2I) generation with embedded textual content, we adopt four complementary perspectives that jointly evaluate: (1) semantic alignment with the prompt, (2) textual accuracy of rendered content, (3) spatial layout consistency, and (4) completeness of prompt coverage.

\textbf{Semantic Alignment.} We report \textit{CLIP Score (CS)} computed using the CLIP to measure the global semantic consistency between the generated image and the input prompt.

\textbf{Textual Accuracy.} We apply PaddleOCRto extract text from generated images, and compare it with prompt-grounded reference spans using the following metrics:
\emph{i}. \textit{Accuracy (Acc)}: Word-level exact match rate.
\emph{ii}. \textit{F1 Score (F1)}: Harmonic mean of word-level precision and recall.
\emph{iii}. \textit{Character Error Rate (CER)}: Normalized edit distance at the character level.

\textbf{Layout Consistency.} We report \textit{Layout IoU (IOU)}, the average Intersection-over-Union between OCR-detected bounding boxes and the ground-truth reference boxes, reflecting the accuracy of the spatial layout.

\textbf{Prompt Coverage.} To evaluate how completely the model renders the semantic content specified in the prompt, we introduce the metric \textit{Prompt Coverage (PC)}:
\[
\text{PC} = \frac{|\mathcal{S}_{\text{matched}}|}{|\mathcal{S}_{\text{total}}|}
\]
where $\mathcal{S}_{\text{total}}$ denotes the set of prompt-specified text spans, and $\mathcal{S}_{\text{matched}}$ is the subset successfully rendered and identified via OCR.
Each target span is explicitly marked in the prompt using quotation marks. A span is considered matched if any OCR-extracted text in the generated image exhibits either an exact or approximate match, determined via fuzzy string comparison with a normalized Levenshtein similarity threshold of 0.8. This metric reflects the model’s ability to accurately render prompt-specified content and serves as an indicator of prompt fidelity.

\section{Experiments}

\subsection{Setup}

\paragraph{Implementation Details.} 
We fine-tune Janus Pro 1B on \DatasetName using 128 V100 GPUs (16 nodes × 8 GPUs) at a $512 \times 512$ resolution and a batch size of 512 for 60k steps. Training uses mixed-precision, DeepSpeed ZeRO-2 with CPU offloading, and AdamW optimization ($\beta_1{=}0.9$, $\beta_2{=}0.95$, weight decay 0.1, $\epsilon{=}1\text{e}{-8}$) with a learning rate of $2\text{e}{-4}$ and linear warm-up for the first 1000 steps.

\begin{table*}[t]

\centering
\footnotesize
\setlength{\tabcolsep}{2pt}
\renewcommand{\arraystretch}{1.1}
\begin{tabular}{l|cccccc|cccccc|cccccc}

\toprule
\textbf{Method} 
& \multicolumn{6}{c|}{\textbf{Eazy}} 
& \multicolumn{6}{c|}{\textbf{Mid}} 
& \multicolumn{6}{c}{\textbf{Hard}} \\
& CS & Acc & F1 & CER &IOU & PC
& CS & Acc & F1 & CER & IOU & PC
& CS & Acc & F1 & CER & IOU & PC\\
\midrule
PixArt-$\Sigma$ & 27.70& 1.06& 0.63 & 82.71& {0.00} & 0.38& 25.72& 0.93& 0.73 & 82.53 & 0.00& 0.15& 19.12& 0.78& 0.68 & 81.86 & 0.00&0.00\\
AnyText & 27.37 & 19.70 & 22.61 & 82.99 & -& - & 26.47 & 3.88 & 6.23 & 88.94 & - & - & - & - & - & - & - & - \\
TextDiffuser-2 & 29.47 & 30.69 & 34.19 & 88.94 & 4.12 & 26.77 & 27.36 & 15.71 & 22.38 & 90.57 & 0.71 & 8.19 & 20.06 & 4.36 & 7.05 & 94.87 & 0.15 & 1.18\\
Janus Pro 7B & 30.33 & 34.87 & 33.80 & 83.35 & 4.79 & 30.42& 28.26 & 14.89 & 17. 57 & 82.45 & 1.60 & 12.58 & 19.17 & 4.87 & 7.22 & 86.94 & 0.19&5.98 \\
SD-3.5 Large & 31.39& 80.60 & 49.72 & 74.55 & 8.61 & 72.24& 29.02 & 76.81 & 58.44  & 63.95 & 6.30 & 65.45 & 17.28 & 58.69 & 45.11&68.57&1.89&38.35 \\
FLUX.1-dev &  {26.93} & 79.35 & 43.98 & 75.71 & 9.17 & 82.00 & 28.97 & 70.18& 64.23 &73.06& 9.24 & 60.29& 19.27& 51.85 & 37.46 &76.00&2.82&60.76\\
\midrule
DALL·E 3 & 26.48& 58.82 & 38.10 & 88.38 & 6.47 & 49.09& 28.60 & 54.45 & 40.48 & 88.43& 4.58& 46.38& 19.70 & 48.07 & 36.98 & 81.45& 1.36&42.35 \\
GPT-4o  & 26.65& 84.45& 50.30 & 73.52 & 18.23 & 84.91& 29.31& 86.63 & 79.09 & 68.51& 15.22 &82.61 & 19.15 & 77.90 & 65.43 & 64.74& 8.44 &83.53 \\
\midrule
\textit{Janus Pro 1B} & 28.71 & 10.85 & 10.00 & 89.86 & 0.80 & 7.22 & 26.75 & 4.48 & 5.52 & 88.72 & 0.27 & 3.33 & 19.79 & 1.46 & 2.24 & 89.95 & 0.00&1.07\\ 
\textit{Janus Pro 1B}$^\dag$ &27.92 & 20.27  & 20.09 & 87.54 &3.80 & 15.02 & 26.53  &  10.25& 13.36 & 86.53  & 1.37  & 7.58 & 19.86  & 2.60 & 4.23 & 91.22 &0.38 & 2.46\\
\quad w/ Text Only$^\dag$& 27.54& 20.46& 20.26 & 86.04 & 3.50& 16.35 & 26.40& 11.78 & 15.22 & 85.04 & 1.76 & 7.88 & 20.21 & 2.53 & 4.15 & 91.05& 0.43 &2.88 \\
\quad w/ Bbox Only$^\dag$ & 28.90 & 34.29& 33.77& 83.83& 7.95 & 30.61 & 27.03 & 17.95 & 23.03 & 82.81& 3.67 & 16.97 & 19.94 & 4.37 & 7.12 & 90.53& 0.55&5.34\\
\quad Pre-Image$^\dag$ & 29.38 & 38.62 & 37.26 & 82.58 & 10.43 & 34.41 & 27.30 & 20.33 & 26.14 & 82.67 & 4.99 & 19.55 & 19.71 & 4.72 & 7.43 & 89.78 & 0.43 &6.30\\
\quad Ours$^\dag$& 29.04 & 34.10& 33.30& 83.02& 7.93 & 30.80& 27.54& 21.90& 27.75 & 81.23& 4.32 & 19.09& 19.74 & 4.53 & 7.33 & 90.57& 0.80&6.09 \\
\bottomrule 
\end{tabular}
\caption{Comparison of text rendering performance across state-of-the-art T2I models and our fine-tuned Janus Pro variants. he best scores are in bold while the second best are in underline. Metrics include CLIP Score (CS) $\uparrow$, OCR Accuracy (Acc.) $\uparrow$, F1 Score (F1.) $\uparrow$,  Character Error Rate (CER) $\downarrow$, Layout IOU (IOU) $\uparrow$ and Prompt Coverage (PC)$\uparrow$. 
$\dag$ means model fine-tuned on \DatasetName\  using $512 \times 512$ resolution.
}
\label{tab:eval_all_subsets}
\end{table*}

\paragraph{Compared Models.}
To evaluate the utility of our dataset and the effectiveness of our training strategy for T2I generation, we fine-tune a lightweight autoregressive model, Janus Pro 1B. We upsample it to 512×512 to better capture fine-grained textual details.
We benchmark our model against six open-source baselines, AnyText,PixArt-$\Sigma$,, TextDiffuser-2,Janus Pro, SD 3.5 Large~\cite{sd3}, and Flux-1-dev~\cite{flux}—as well as two proprietary systems, DALL·E 3 and GPT-4o~\cite{gpt4o}.

\subsection{Main Results}
Table~\ref{tab:eval_all_subsets} summarizes the evaluation results across baselines and variants of our model. 
Compared to the vanilla Janus Pro 1B model, which is not fine-tuned on our dataset, the fully supervised version (\textbf{Ours}) shows significant improvements across all subsets. This validating the Effectiveness of Our Dataset.
Proprietary models such as GPT-4o exhibit strong prompt understanding capabilities, achieving the highest prompt coverage across all difficulty levels (e.g., 83.53 on hard subset).
Our method stands out in robustness under increasing complexity. While baselines TextDiffuser-2 and Janus Pro 7B suffer major performance drops on the hard subset, our model maintains performance, demonstrating better compositional understanding and spatial fidelity when handling complex, multi-span prompts.

\subsection{Ablation Studies}
We conduct ablation studies to investigate two core aspects of our training strategy: the effect of supervision tokens and the influence of token ordering.

\paragraph{Effect of Supervision Tokens.}  

To assess the contribution of different supervision signals, we conduct an ablation study based on the Janus Pro 1B model, trained at $512\times512$ resolution using our dataset. We compare four variants:

\emph{i.} \textbf{Vanilla Fine-tune}: Fine-tuned without any additional supervision tokens.
\emph{ii.} \textit{Text Only}: Fine-tuned using OCR word tokens appended after image tokens during training.
\emph{iii.} \textit{BBox Only}: Fine-tuned using bounding box tokens only, without word-level textual supervision.
\emph{iv.} \textit{Text + BBox (Ours)}: Fine-tuned using both word and bbox tokens jointly as a unified sequence.

As shown in Table~\ref{tab:eval_all_subsets}, our ablation results reveal three key insights. First, adding only text token supervision yields a modest improvement over vanilla fine-tuning. This suggests that although prompts already contain textual cues, explicitly supervising the generated text provides slight additional guidance. Second, using only bounding box tokens as supervision leads to a substantial performance gain. This implies that spatial layout information offers a more distinct and complementary signal for guiding text placement. Finally, our full method—combining both text and bounding box tokens—performs similarly to the Bbox-only variant on the Easy subset, but exhibits significantly better results on the Mid and Hard subsets. This indicates that \textit{with growing layout complexity and task difficulty, the joint supervision strategy proves to be more effective and robust}.

\paragraph{Effect of Token Order.}
We investigate how the placement of supervision tokens within the target sequence influences learning, comparing two configurations: \emph{i}. \textit{Pre-Image}, where supervision tokens precede image tokens, and \emph{ii}. \textit{Post-Image (Ours)}, where they follow image tokens.

As shown in Table~\ref{tab:eval_all_subsets}, the Pre-Image variant shows a slight advantage on the Easy subset (PC: 34.41 vs. 30.80; IOU: 10.43 vs. 7.93), but this benefit diminishes under higher complexity. On the Hard subset, both achieve similar PC (6.30 vs. 6.09), while Post-Image yields better layout alignment (IOU: 0.80 vs. 0.43).

These findings suggest that earlier exposure to layout tokens can help the model form preliminary layout plans. However, the Post-Image strategy offers a fully inference-consistent design—requiring no changes to the decoding process—while maintaining stable performance across all difficulty levels. This makes Post-Image a more practical and robust default choice.

\subsection{Qualitative Analysis}

To study the impact of our proposed dataset and training strategy, we conduct a series of qualitative analyses focusing on different supervision settings and model variants. Specifically, we present four categories of visual comparisons: (1) before and after fine-tuning on our dataset to verify the overall effectiveness of layout-aware supervision; (2) a detailed comparison among Text Only, Bbox Only, and Text\&Bbox supervision to reveal their respective behaviors and limitations. 
More visualizations are in the supplementary.
\paragraph{Before vs. After Fine-tuning.} 
\begin{figure}[t]
    \centering
    \includegraphics[width=0.9\linewidth]{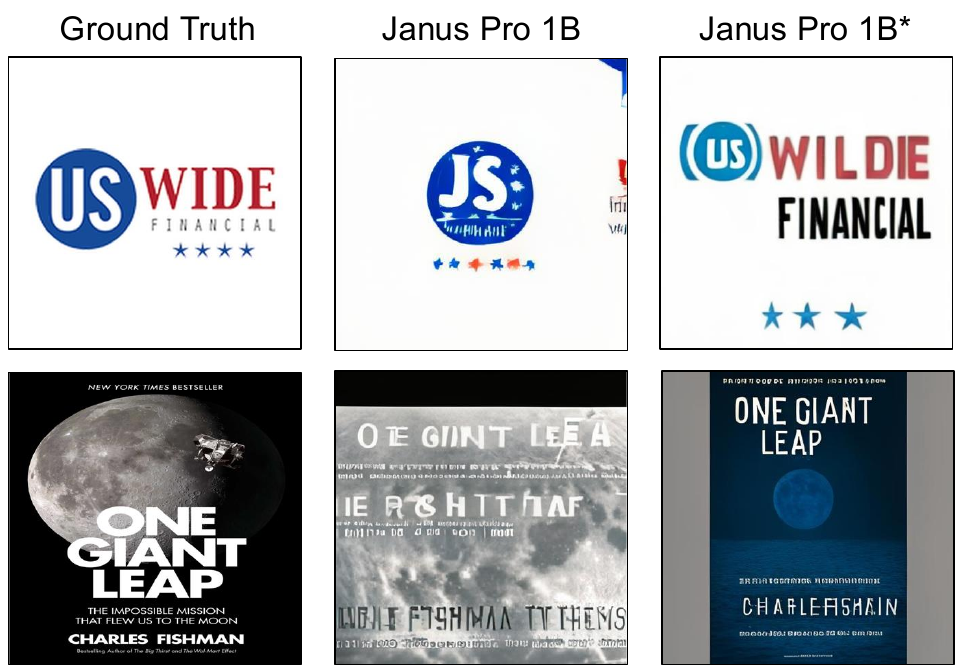}
    \caption{ Qualitative comparison of generation results before and after fine-tuning. 
    } 
    \label{fig:data_effect}
\end{figure}
We compare the Janus Pro 1B model and fine-tuned version to verify the impact of our dataset. 
Fine-tuning yields significant gains in layout consistency and textual accuracy, as illustrated in Figure~\ref{fig:data_effect}.

\paragraph{Text Only vs. BBox Only vs. Text \& Bbox.}
\begin{figure}[t]
    \centering
    \includegraphics[width=.9\linewidth]{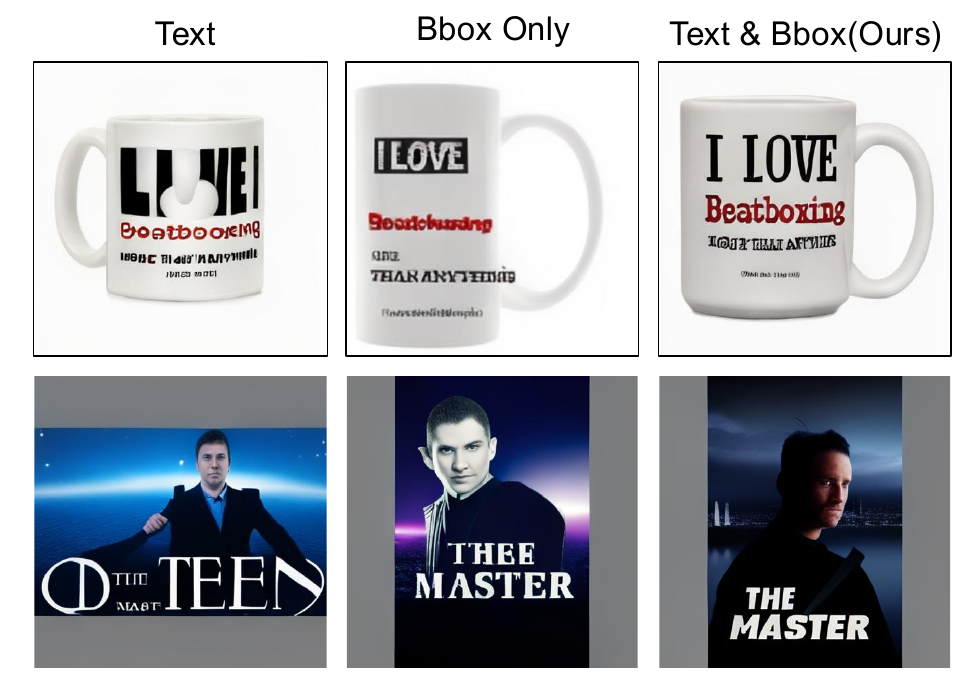}
    \caption{ Qualitative comparison of different supervision strategies: text-only, bbox-only, and text \& bbox. 
    } 
    \label{fig:bbox_effect}
\end{figure}
We further compare the effects of different supervision signals.  As shown in Figure~\ref{fig:bbox_effect}, using text supervision only often leads to layout inconsistencies or even failure to render text properly. In contrast, using bounding boxes only improves spatial layout but may produce inaccurate or extraneous text—e.g., in the second row, second column, an extra letter "E" is hallucinated in "THEE MASTER". Our mothod, which combines both text and bounding box supervision, results in significantly better alignment between prompt and image content, achieving accurate and well-positioned text rendering across diverse layouts. More visual results and analysis can be found in the supplementary.

\section{Conclusion}

In this work, we tackle the underexplored problem of prompt-grounded, layout-aware text rendering in text-to-image (T2I) generation. We present \DatasetName, a large-scale dataset with prompt-aligned annotations, and introduce a lightweight training strategy that injects layout supervision into autoregressive models without altering their architecture or inference. Experiments across difficulty-aware benchmarks show improved spatial fidelity and prompt alignment. 
While our dataset provides spatial annotations for prompt-grounded text rendering, it currently lacks fine-grained typographic details such as font and color. In future work, we plan to enrich these annotations to better support high-fidelity and stylistic T2I generation tasks.

\bibliography{aaai2026}

\end{document}